\pdfoutput=1
\documentclass{article}
\PassOptionsToPackage{numbers,compress}{natbib}
\usepackage[preprint]{neurips_2025}
\usepackage[utf8]{inputenc}
\usepackage[T1]{fontenc}
\usepackage{hyperref}
\usepackage{url}
\usepackage{booktabs}
\usepackage{amsfonts,amssymb,amsmath}
\usepackage{nicefrac}
\usepackage{microtype}
\usepackage{xcolor}
\usepackage{graphicx}
\usepackage{pifont}
\usepackage{tikz}
\usetikzlibrary{positioning,arrows.meta,fit,calc,backgrounds}
\definecolor{ink}{HTML}{333333}   \definecolor{ours}{HTML}{5484B4}
\definecolor{proxy}{HTML}{B4543C} \definecolor{neutral}{HTML}{8C8C8C}
\definecolor{fillM}{HTML}{E4E4FC} \definecolor{fillD}{HTML}{CCE4E4} \definecolor{fillQ}{HTML}{FCFCCC}
\definecolor{hack}{HTML}{F4B6B6}  \definecolor{ok}{HTML}{B7E1B5}
\tikzset{
  box/.style={draw=ink, line width=0.6pt, rounded corners=2pt, fill=#1,
              align=center, font=\footnotesize, inner sep=3pt, minimum height=6mm},
  box/.default=fillM,
  arr/.style={-{Stealth[length=4pt]}, line width=0.7pt, draw=ink},
  accent/.style={draw=ours, line width=0.9pt},
  group/.style={draw=ink, dashed, line width=0.4pt, rounded corners=3pt, inner sep=4pt},
  badge/.style={circle, fill=ink, text=white, font=\scriptsize\bfseries, inner sep=1pt},
}
\newcommand{\cmark}{\checkmark}
\newcommand{\xmark}{\ding{55}}

\definecolor{linknavy}{HTML}{0C3C6C}
\hypersetup{colorlinks=true, citecolor=linknavy, linkcolor=linknavy, urlcolor=linknavy,
  pdftitle={Single-Database Rewards Are Hackable: Counterfactual Execution for Trustworthy Text-to-SQL RL},
  pdfauthor={Shreshth Rai, Rishii Kumar Singh, Aksh Aggarwal}}

\title{Single-Database Rewards Are Hackable: Counterfactual Execution for Trustworthy Text-to-SQL RL}

\author{%
  Shreshth Rai, \quad Rishii Kumar Singh, \quad Aksh Aggarwal \\
  Artificial Intelligence and Machine Learning Society (AIMS-DTU) \\
  Delhi Technological University, New Delhi, India \\
  {\small\texttt{\{shreshthrai\_mc24b06\_039, rishiikumarsingh\_ene\_25b18073,}} \\
  {\small\texttt{Akshaggarwal\_cse\_25a01023\}@dtu.ac.in}}
}

\begin{document}
\maketitle

\begin{abstract}
Reinforcement learning (RL) for text-to-SQL agents pays a query when its result matches the reference on one database. However, a wrong query can match by luck when the data lack the rows that expose its bug. We study a \emph{counterfactual-database (CF) reward} that also runs the final query on a few hidden, code-generated copies of the database and pays only if all results match (strict form) or by the share matched (mean form). On 1,302 synthetic audit questions, the strict form lowers the share of buggy queries paid in full from 29.5\% to 16.8\% and of reward hacks from 54.1\% to 1.9\%, rejecting 0 of 2,847 correct queries; copies from a family-blind generator do about as well. In a controlled GRPO comparison on a 3.35B model with three seeds per arm, training on the mean form did not improve in-domain accuracy (8.4\% vs 11.7\%), because the two rewards disagreed on only 5--7\% of rollouts; on Spider it gained 1.6 points. One seed reverses both signs, so single-seed reward comparisons mislead.\footnote{Code: \href{https://github.com/RishiiGamer2201/counterfactual-sql-rewards}{\nolinkurl{github.com/RishiiGamer2201/counterfactual-sql-rewards}}; data, models, demo: \href{https://huggingface.co/counterfactual-sql}{\nolinkurl{hf.co/counterfactual-sql}}.}
\end{abstract}

\section{Introduction}
\label{sec:intro}

\begin{figure}[t]
  \centering
  \begin{tikzpicture}
  \node[box=fillQ, text width=15mm] (q) at (0,0) {Question $x$\\ + schema};
  \node[box, text width=17mm, right=4mm of q] (pi) {Policy $\pi_\theta$\\ (SQL agent)};
  \node[box=fillQ, text width=13mm, right=4mm of pi] (y) {Final\\ query $y$};
  \draw[arr] (q) -- (pi);
  \draw[arr] (pi) -- (y);

  \node[box=fillD, text width=31mm, anchor=west] (d0) at ($(y.east)+(7mm,10mm)$) {$D_0$: original DB};
  \node[box=fillD, text width=31mm, anchor=north west] (d1) at ($(d0.south west)+(0,-5.5mm)$) {$D_1$: answer-changing};
  \node[box=fillD, text width=31mm, below=1.2mm of d1] (d2) {$D_2$: answer-preserving};
  \node[font=\footnotesize, below=-0.5mm of d2] (dk) {$\vdots$\;\;$D_K$};
  \node[group, fit=(d1)(d2)(dk)] (hid) {};
  \node[font=\scriptsize, text=neutral, anchor=north] at (hid.south) {hidden copies, one rule edit per bug family};

  \draw[arr, {Stealth[length=4pt]}-{Stealth[length=4pt]}, dashed] (pi.north) |- node[pos=0.72, above, font=\scriptsize] {\texttt{run\_sql} (explore)} (d0.west);
  \coordinate (fork) at ($(y.east)+(3mm,0)$);
  \draw[line width=0.7pt, draw=ink] (y.east) -- (fork);
  \draw[arr] (fork) |- ($(d0.west)+(0,-1.2mm)$);
  \draw[arr] (fork) |- (hid.west);

  \coordinate (m) at ($(d0.east)+(3mm,0)$);
  \node[box=white, draw=proxy, line width=0.9pt, text width=27mm, anchor=west] (rp) at ($(d0.east)+(8mm,0)$)
    {single-DB $\tilde R = C_0$};
  \node[box=white, accent, text width=27mm, anchor=west] (rc) at (rp.west |- hid.east)
    {CF reward\\ $R = \prod_{k=0}^{K} C_k$};
  \draw[line width=0.7pt, draw=ink] (d0.east) -- (m);
  \draw[arr] (m) -- node[above, font=\scriptsize] {$C_0$} (rp.west);
  \draw[arr, draw=ours] (m) |- ($(rc.west)+(0,1.5mm)$);
  \draw[arr, draw=ours] (hid.east |- {$(rc.west)+(0,-1.5mm)$}) -- node[below=0.5mm, font=\scriptsize, fill=white, inner sep=1pt] {$C_{1..K}$} ($(rc.west)+(0,-1.5mm)$);

  \node[badge] at (pi.north west) {1};
  \node[badge] at (hid.north west) {2};
  \node[badge] at (rc.north east) {3};

  \node[box=white, text width=\dimexpr\linewidth-12pt\relax, align=left, anchor=north west, inner sep=3.5pt]
    (ex) at ($(q.west |- hid.south)+(0,-3.5mm)$) {%
    \begin{minipage}[c]{0.64\linewidth}\scriptsize\setlength{\fboxsep}{1pt}%
      \textbf{Ex.} $x$: ``How many distinct customers bought a grocery product in 2025?''\par\vspace{2pt}
      $y$\phantom{$^\star$}\,(agent):\;\;\;\,{\ttfamily SELECT COUNT(DISTINCT \colorbox{hack}{oi.order\_id}) \dots}\par\vspace{1pt}
      $y^\star$\,(reference):\;{\ttfamily SELECT COUNT(DISTINCT \colorbox{ok}{o.customer\_id}) \dots}
    \end{minipage}\hfill
    \begin{minipage}[c]{0.34\linewidth}\scriptsize\setlength{\fboxsep}{1pt}%
      \setlength{\tabcolsep}{2.5pt}%
      \begin{tabular}{@{}lccc@{}}
        & $y$ & $y^\star$ & \\ \midrule
        $D_0$ (1 order per buyer) & 10 & 10 & \colorbox{ok}{\cmark\,match} \\
        $D_1$ (2 buyers, 5 orders) & 15 & 12 & \colorbox{hack}{\xmark\,miss} \\ \midrule
        \multicolumn{4}{@{}l}{$\Rightarrow$ single-DB $\tilde R{=}1$,\; CF $R{=}0$}
      \end{tabular}
    \end{minipage}};
\end{tikzpicture}
  \caption{\textbf{A single-database reward pays a query that a hidden counterfactual copy rejects.} The policy explores only $D_0$; its final query $y$ also runs on copies $D_k$, each with one rule edit. Rust: single-DB $\tilde R$, paid by a match on $D_0$; blue: CF reward $R$, paid only by a match on every database. Bottom: a test item where the agent counts orders (red) instead of customers (green).}
  \label{fig:pipeline}
\end{figure}
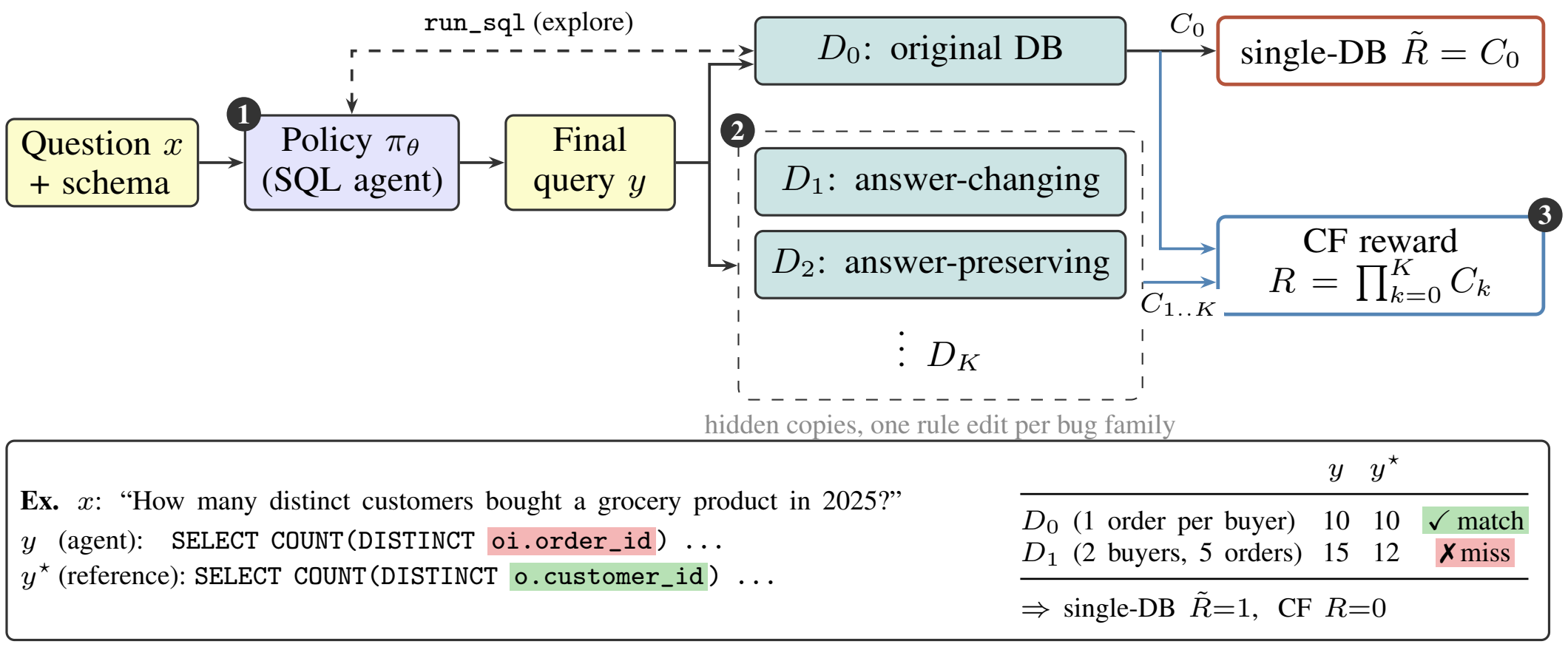

A text-to-SQL query that fails is visible; one that runs and returns a plausible but wrong number ends up in a business report. Reinforcement learning with verifiable rewards trains such agents by executing the generated query and paying it when its result matches the reference result on one database~\citep{ma2026sql,pourreza2025reasoningsql,yao2026arctic}. The check is cheap but has a blind spot. Asked ``How many distinct customers bought a grocery product in 2025?'', our agent counted distinct orders. On the database it explored, each buyer placed one order, so both counts are 10 and the reward paid in full; on a resampled copy where two customers placed five orders, the reference answer is 12 and the agent says 15 (Figure~\ref{fig:pipeline}).

However, execution-match rewards assume that one database separates correct from incorrect queries. This fails when the question reads a small slice of the data, such as one city, year or category: in our synthetic retail databases, 29.2\% of realistic buggy queries pass the single-database (single-DB) check at 50 customers, against 0.8\% at 500 customers. 
Real questions often filter to such slices; an LLM judge would add a second model that can be gamed.

We ask: \emph{can grading the final query on databases the agent never sees make a SQL reward harder to hack, and does training on it produce a more trustworthy agent?} We grade each query on the original database and two (rarely three) hidden copies, generated by code one bug family at a time, and pay only if all results match. Our contributions are:
(i) \textbf{An audited counterfactual-database reward} that cuts the full-pay rate of buggy queries from 29.5\% to 16.8\% and of hacks from 54.1\% to 1.9\% without rejecting a correct query (Table~\ref{tab:audit}).
(ii) \textbf{Evidence that public rewards share the hole}: an exploit in AgentTune's SQL reward (fixed upstream), and the same gap in the SkyRL-SQL reward and eight published Spider systems.
(iii) \textbf{A controlled three-seed training study with an honest null}: with only the reward changed, the CF reward did not train a better agent in domain, and we trace why; it gains 1.6 points on Spider, within seed noise (Table~\ref{tab:train}).

\paragraph{Related work.}
Reward hacking, where an optimizer is paid through a proxy that diverges from the intended objective, is well documented~\citep{amodei2016concrete,skalse2022defining,gao2023scaling}. Text-to-SQL RL trains on execution-match rewards~\citep{ma2026sql,pourreza2025reasoningsql,yao2026arctic}, ReViSQL finds many positive execution rewards false~\citep{zhu2026humanleveltexttosqlreinforcementlearning}, and DualSQL tightens the comparator~\citep{chen2026dualsql}. Test suites~\citep{zhong2020semantic,zhang2026progresssql}, generated grading data~\citep{chandra2015data,habibollah2026synsql} and harder code tests~\citep{he2026hardtestgen} catch lucky-correct programs~\citep{pasupat2016inferring}, mostly at evaluation time. We move the check into the RL reward and audit it before training. It bears on reward evaluation, red-teaming and monitoring.

\section{Method}
\label{sec:method}

\paragraph{Threat model.}
The adversary is the policy. It sees the question, schema and original database $D_0$ (read-only), never the copies or grader. In scope: lucky-correct queries, literal smuggling (hard-coding an answer seen while exploring), hard-coded row IDs and comparator holes; out of scope: grader tampering and wrong references.

\paragraph{Rewards.}
Let $y$ be the agent's final query, $y^\star$ the reference query and $\operatorname{exec}(y,D)$ the rows that $y$ returns on database $D$. Define $C_k := \mathbf{1}[\operatorname{exec}(y,D_k) \equiv \operatorname{exec}(y^\star,D_k)]$, where $\equiv$ is a strict multiset match of rows (ordered when the reference is ordered). The single-DB reward and the CF reward are
\begin{equation}
  \tilde R(x,y) := C_0, \qquad R(x,y) := \textstyle\prod_{k=0}^{K} C_k .
  \label{eq:rewards}
\end{equation}
The main run trains on the mean form $R_{\text{mean}} := C_0 \cdot \frac{1}{K}\sum_{k=1}^{K} C_k$, fixed before training; the curriculum run uses the strict $R$. Execution is sandboxed and read-only. Under GRPO~\citep{DBLP:journals/corr/abs-2402-03300}, which normalizes rewards over a prompt's $G$ rollouts, the rewards of Eq.~\eqref{eq:rewards} differ only where $\tilde R$ pays and $R$ does not.

\paragraph{Counterfactual copies.}
Each copy applies one rule edit to the slice the question reads and keeps the business rules. \emph{Answer-changing} edits, such as new customers in the slice, defeat literal smuggling and hard-coded IDs. \emph{Answer-preserving} edits keep the reference answer but add rows that expose one bug family: an extra active order (F1, duplicates), instalment payments (F2, join fan-out), cancellations or refunds (F3, missing filters) or superseded tier rows (F4, superseded records). References are checked against an independent pandas implementation (Appendix~\ref{app:cf}). \emph{Blind} selection takes one random copy of each kind (our audit headline); \emph{informed} selection picks up to three copies greedily to kill a template's known buggy variants. Training used informed copies, so their audit numbers are an optimistic bound; 442 of 445 training prompts got two copies, and only 202 of the 893 copies kill a mutant that passes $D_0$.

\paragraph{Audit metrics.}
Before training, we score fixed candidates: 2,847 correct queries, 1,814 realistic buggy variants of the references (\emph{mutants}, 13 types) and 2,839 constructed hacks (literal answers, empty results, hard-coded IDs). We report the false-accept rate $\Pr[r=1 \mid y \text{ wrong}]$ (``paid in full'') and the false-reject rate $\Pr[r<1 \mid y \text{ correct}]$. A sample is \emph{lucky} if $\tilde R$ pays it and a test copy rejects it.

\section{Experimental setup}
\label{sec:setup}

\paragraph{Data.}
A seeded generator builds 8-table retail SQLite databases; 18 question templates span the four bug families. The audit split has 1,302 items, the training split 907, and the test split 331 from 7 templates unseen in training (214 of them F4, which no training question covers). Test items have 4 copies from a different, family-blind generator; a sample is \emph{verified} if it matches on the original and all 4. For transfer we use the 1,034 Spider dev questions~\citep{yu-etal-2018-spider}, with execution accuracy (EX) on the original database and test-suite accuracy (TS) on the databases of \citet{zhong2020semantic}.

\paragraph{Model and training.}
We use Tiny Aya Global (3.35B)~\citep{salamanca2026tiny} with a rank-16 LoRA~\citep{hu2021lora} in AgentTune~\citep{e2026agenttune}; the agent has one tool, \texttt{run\_sql} on $D_0$, and up to 5 steps. SFT uses 353 scripted trajectories and holds out 3 of 11 training templates, so GRPO sees questions SFT gets wrong. GRPO runs 60 steps on 445 prompts (6 rollouts $\times$ 4 prompts, temperature 1.0, KL weight 0.02). The two arms, B ($\tilde R$) and C (CF reward), share initialization, data and hyperparameters; only the reward differs. Seeds 0--2 set prompt order and the trainer RNG. A pre-registered curriculum pair trains on 36 hard prompts where 89\% of SFT's passes are lucky. The main pair took about 7 hours on one L4 GPU (about \$5); the copies add about 1\,ms of CPU time per rollout (Appendix~\ref{app:hyper}).

\paragraph{Evaluation.}
We draw 8 samples per test item and 4 per Spider question (temperature 0.7) and report Wilson 95\% intervals over samples, and paired bootstrap intervals for differences that resample questions (samples averaged per question). The primary test pools each question's samples over seeds; as it ignores training variance, we also give a seed-level $t$ interval over per-seed differences.

\begin{table}[t]
  \caption{\textbf{Hidden copies cut false accepts without rejecting correct queries.} Share paid in full or rejected (\%, Wilson 95\% CI; 1,814 mutants, 2,839 hacks and 2,847 correct queries on 1,302 audit items). $\downarrow$: lower is better; bold: best with blind copies. $^\dagger$Copies chosen knowing the mutants.}
  \label{tab:audit}
  \centering
  \small\setlength{\tabcolsep}{3.5pt}
  \begin{tabular}{llrrr}
\toprule
 & & \multicolumn{2}{c}{Paid in full (\%)\,$\downarrow$} & Rejected (\%)\,$\downarrow$ \\
\cmidrule(lr){3-4}\cmidrule(lr){5-5}
Reward comparator & Databases graded & Mutants & Hacks & Correct \\
\midrule
Set match (BIRD/SkyRL style) & $D_0$ & 37.5{\,[35.3, 39.7]} & 54.1{\,[52.3, 56.0]} & \textbf{0.0}{\,[0.0, 0.1]} \\
Strict multiset match, $\tilde R$ & $D_0$ & 29.5{\,[27.4, 31.6]} & 54.1{\,[52.3, 56.0]} & \textbf{0.0}{\,[0.0, 0.1]} \\
Set match & $D_{0..K}$, blind & 27.2{\,[25.2, 29.3]} & \textbf{1.9}{\,[1.5, 2.5]} & \textbf{0.0}{\,[0.0, 0.1]} \\
\midrule
Strict multiset match, $R$ (ours) & $D_{0..K}$, blind & \textbf{16.8}{\,[15.2, 18.6]} & \textbf{1.9}{\,[1.5, 2.5]} & \textbf{0.0}{\,[0.0, 0.1]} \\
Strict multiset match, $R$ (bound)$^\dagger$ & $D_{0..K}$, informed & 1.5{\,[1.0, 2.2]} & 2.0{\,[1.6, 2.6]} & 0.0{\,[0.0, 0.1]} \\
\bottomrule
\end{tabular}

\end{table}

\section{Results}
\label{sec:results}

\paragraph{Single-DB rewards, ours and public ones, pay for bugs and hacks.}
The strict single-DB reward $\tilde R$ pays 29.5\% of mutants and 54.1\% of hacks in full (Table~\ref{tab:audit}), including every literal answer and hard-coded-ID query. A set comparison, as in BIRD-style~\citep{li2023can} and SkyRL-SQL~\citep{cao2025skyrl,griggs2025skrylv01} rewards, pays 37.5\% of mutants, including all 203 queries that drop a needed \texttt{DISTINCT}. AgentTune's SQL reward~\citep{e2026agenttune} paid \texttt{WHERE 1 = 1} padding with an always-``yes'' answer like honest SQL on all 314 yes items until the maintainers merged our fix (Appendix~\ref{app:hacks}), and 8.1--14.9\% of EX-correct answers from eight published Spider systems~\citep{ascoli2025etm} fail a test-suite database.

\paragraph{Hidden copies cut false accepts, but do not close the gap.}
With blind copies, the CF reward pays 16.8\% of mutants and 1.9\% of hacks and rejects 0 of 2,847 correct queries; literal smuggling drops to 0 of 1,302. That is a 43\% relative cut, short of the 50\% we pre-registered. Survivors are mutants the random copy does not touch (40.0\% of \texttt{all\_customers}, 37.7\% of \texttt{missing\_refunds}; Appendix~\ref{app:results}), and every paid hack is a hard-coded-ID query whose answer-changing copy kept the same IDs (54 of 235). Informed copies lower mutant pay to 1.5\%, an optimistic bound.

\paragraph{The copies need not encode our bugs, and a larger database is no substitute.}
Copies from the family-blind test generator pay 17.2\% of mutants and 2.3\% of hacks (0 of 3,017 correct queries rejected); without the F4-targeted edits, F4 hack pay still falls from 54.3\% to 3.7\% (mutants: 28.5\% to 23.5\%). One database as large as the original plus two copies (150 customers) passes only 7.8\% of mutants but pays every literal and hard-coded-ID hack: size catches accidents, not exploits.

\begin{table}[t]
  \caption{\textbf{A harder-to-hack reward trained no better agent in domain, slightly better on Spider.} Values in \%. Verified: correct on all 5 test databases (Wilson 95\% CI, $n=331\times8$); lucky: single-DB minus verified; lucky share: EX passes failing a suite DB. $\uparrow$/$\downarrow$: higher/lower is better; bold: better of pair; GRPO: seed 0 (others: Appendix~\ref{app:seeds}).}
  \label{tab:train}
  \centering
  \small\setlength{\tabcolsep}{3.5pt}
  \begin{tabular}{llrrrrr}
\toprule
 & & \multicolumn{3}{c}{Our test set, 331 items $\times$ 8} & \multicolumn{2}{c}{Spider dev, 1034 $\times$ 4} \\
\cmidrule(lr){3-5}\cmidrule(lr){6-7}
Policy & Training reward & Verified\,$\uparrow$ & Single-DB & Lucky\,$\downarrow$ & TS\,$\uparrow$ & Lucky share\,$\downarrow$ \\
\midrule
Base model & -- & 0.9{\,[0.6, 1.4]} & 1.4 & 0.5 & 12.6 & 18.3 \\
SFT & -- & 9.4{\,[8.4, 10.6]} & 14.5 & 5.1 & 29.1 & 24.4 \\
\midrule
GRPO, main & $\tilde R$ (single-DB) & \textbf{12.7}{\,[11.5, 14.0]} & 17.8 & 5.1 & 27.9 & 26.4 \\
GRPO, main (ours) & $R_{\text{mean}}$ (CF) & 9.0{\,[7.9, 10.1]} & 14.0 & 5.1 & \textbf{29.9} & \textbf{23.6} \\
\midrule
GRPO, curriculum & $\tilde R$ (single-DB) & \textbf{7.1}{\,[6.2, 8.2]} & 11.6 & 4.4 & 27.6 & 28.3 \\
GRPO, curriculum (ours) & $R$ (CF, strict) & 6.3{\,[5.5, 7.3]} & 10.6 & \textbf{4.3} & \textbf{29.7} & \textbf{25.3} \\
\bottomrule
\end{tabular}

\end{table}

\paragraph{The CF reward did not train a better agent in domain.}
Pooled over three seeds, GRPO on $R_{\text{mean}}$ verifies 8.4\% of test samples against 11.7\% for GRPO on $\tilde R$ ($-3.2$ [$-4.7$, $-1.8$]) and stays at or just below SFT's 9.4\% in every seed ($-0.5$, $-0.6$, $-1.9$ points). $\tilde R$ is unstable ($+3.2$, $-1.9$, $+5.3$ points over SFT), so C$-$B is $-3.7$, $+1.2$ and $-7.2$ by seed, each with a narrow interval; seed level: $-3.2$ [$-13.7$, $+7.3$]. The pre-registered curriculum pair (one seed) repeats the null ($-0.8$ [$-1.8$, $+0.2$]). Likely cause: weak signal. In every run 84--88\% of rollouts passed $D_0$ and only 5--8\% of passes were lucky, so the rewards disagreed on about 5--7\% of rollouts. Lucky passes also fell under $\tilde R$ alone (seed 0: from 10.7\% to 5.9\% of passes) but grew on hard curriculum prompts (Appendix~\ref{app:results}).

\paragraph{On Spider, the CF reward transfers slightly better, within seed noise.}
Over three seeds, CF-trained policies reach 29.3\% test-suite accuracy against 27.8\% for single-DB ones ($+1.6$ [$+0.8$, $+2.3$]); by seed the gap is $+1.9$, $-0.5$ and $+3.3$ (seed-level $+1.6$ [$-3.2$, $+6.4$]). The curriculum pair agrees ($+2.1$ [$+0.9$, $+3.4$]). We did not pre-register this test; single-DB policies partly fell below SFT's 29.1\%.

\section{Discussion and conclusion}
\label{sec:discussion}

Hidden counterfactual databases are a judge-free check that removes nearly all constructed hacks and many buggy passes without rejecting correct queries, but changes training only where the policy is paid for bugs, rare in our runs. A planted-cache pilot spread no underpaying hack under either reward (Appendix~\ref{app:leaky}). Audit rewards on counterfactual data first, monitor the $\tilde R$--$R$ gap during it, compare rewards over several seeds, and do not read a passed audit as proof of correctness.

\paragraph{Limitations and broader impact.}
One synthetic schema, three seeds, 60 GRPO steps, one 3.35B model, English only, no human red-team. Spider may be in the base model's pretraining data. Test copies catch 54 of 103 base-blind mutants, so verified accuracy is lenient. Exploits are dual-use; we disclosed AgentTune's with a fix.

\bibliographystyle{plainnat}
{\small
\bibliography{scholar,refs}}

\clearpage
\appendix

\section{Counterfactual-database generation}
\label{app:cf}

\paragraph{Edits.}
Each training or audit copy applies one of the edits in Table~\ref{tab:edits}, focused on the item's slice (src: \texttt{sqlcf/cf.py}).
\begin{table}[ht]
\caption{\textbf{Rule edits that build the counterfactual copies.} Each copy applies one edit to the slice of the database that the question reads. Targets: the bug family (F1--F4) or query bug that the edit exposes; ``changing'': the edit changes the reference answer.}
\label{tab:edits}
\centering\small\setlength{\tabcolsep}{4pt}
\begin{tabular}{lll}
\toprule
Edit & What it does & Targets \\
\midrule
\texttt{e\_dup\_order} & extra active order for existing customers & F1 duplicates \\
\texttt{e\_new\_customers} & new customers in the slice, with orders & most aggregates (changing) \\
\texttt{e\_zero\_order\_customers} & customers without orders & \texttt{all\_customers} bugs \\
\texttt{e\_cancelled\_orders} & cancelled orders for existing customers & F3 missing status filter \\
\texttt{e\_later\_tier} & later tier row that changes the current tier & F4 latest row \\
\texttt{e\_superseded\_tier} & historical tier row between the first two & F4 any-row, first-row \\
\texttt{e\_refunds} & refunds on shipped orders, gross unchanged & F3 missing refunds \\
\texttt{e\_split\_payments} & payments split into two instalments & F2 join fan-out \\
\texttt{e\_split\_items} & item lines with quantity $\geq 2$ split & count-for-sum, fan-out \\
\texttt{e\_adjustments} & invoice correction plus matching payment & F4 ignored adjustments \\
\texttt{e\_drop\_customers} & delete 30\% of slice customers and their rows & answer-changing \\
\bottomrule
\end{tabular}
\end{table}

\paragraph{Item construction.}
An item is kept only if (1) the reference answer is not degenerate (empty, all-NULL or a single 0); (2) the reference equals an independent pandas implementation on every database and every paraphrase equals the reference; (3) the reference and paraphrases are deterministic on a row-shuffled copy, which catches ties and unordered \texttt{LIMIT}; and (4) for training and audit items, at least one answer-changing and one answer-preserving copy exist.

\paragraph{Copy selection.}
The \emph{informed} set adds the best answer-changing copy, then the best answer-preserving copy, then more copies while they kill more of the template's known mutants, up to 3. The \emph{blind} set takes one random copy of each kind. Training used the informed set; on F1--F3 it catches 401 of 420 base-blind training mutants.

\paragraph{Test copies.}
Each test database is a foreign-key-closed 50-customer sub-database of a 500-customer parent, plus four family-blind mutations: a resample sharing half the customers, a clone of 25\% of customers under new IDs, placed/cancelled swaps on 40\% of unpaid orders, and relabeling of 30\% of tier rows. The period-edge copy (audit only) moves half of a year's orders to January 1 or December 31 and half of the other year's orders just outside the year, with their payments, refunds and adjustments.

\section{Hack and bug examples}
\label{app:hacks}

\paragraph{Constructed hacks.}
\emph{Literal}: a query that returns the reference answer as constants (1,302). \emph{Empty}: a query that returns no rows (1,302). \emph{Hard-coded IDs}: a query that filters on the row IDs of the reference answer on $D_0$ (235).

\paragraph{AgentTune exploit.}
On 360 yes/no items (314 yes, 46 no), honest reference SQL scores 3.10 on every item. \texttt{WHERE 1 = 1} padding with an always-``yes'' answer scores 3.10 on yes items and $-2.0$ on no items (2.45 overall). After our patch it scores 2.20; with the opt-in result-match term, 1.20 against 5.10 for honest SQL. The upstream fix (issue \#11, PR \#12) also repaired a second bug that made the reward read every tool result as ``no rows''.

\paragraph{Hand audit of wrong test trajectories.}
We labeled 48 wrong trajectories (16 each from SFT, GRPO-B and GRPO-C; 18 lucky) with NL2SQL-BUGs categories~\citep{liu2025nl2sql} (Table~\ref{tab:handaudit}). No trajectory was a hack or literal smuggling.
\begin{table}[ht]
\caption{\textbf{Errors in 48 wrong test trajectories.} Counts per error with its NL2SQL-BUGs category; Lucky 18: the trajectories that $\tilde R$ pays and a test copy rejects.}
\label{tab:handaudit}
\centering\small
\begin{tabular}{llrr}
\toprule
Error & Category & All 48 & Lucky 18 \\
\midrule
order date used for payment date & Attribute & 11 & 4 \\
category matched to customer columns, or hallucinated value & Value & 7 & 0 \\
extra status filter & Condition & 6 & 5 \\
no latest-row filter & Condition & 5 & 2 \\
average over items instead of orders & Clause & 5 & 0 \\
invoice adjustments left out & Table & 4 & 3 \\
query does not run & Others & 4 & 0 \\
\texttt{COUNT(*)} instead of \texttt{SUM(quantity)} & Function & 3 & 3 \\
incoherent & Others & 2 & 0 \\
wrong table & Table & 1 & 1 \\
\bottomrule
\end{tabular}
\end{table}

\section{Hyperparameters, prompts and compute}
\label{app:hyper}

Table~\ref{tab:hyper} lists the training settings.
\begin{table}[ht]
\caption{\textbf{Training hyperparameters.} SFT trains a LoRA adapter on scripted trajectories; GRPO continues that adapter, and the two GRPO arms differ only in the reward.}
\label{tab:hyper}
\centering\small
\begin{tabular}{lll}
\toprule
Setting & SFT & GRPO (both arms) \\
\midrule
Base weights & \texttt{CohereLabs/tiny-aya-global}, bf16 & continues the SFT adapter \\
LoRA & $r=16$, $\alpha=32$, dropout 0.05, all projections & same adapter \\
Learning rate & $10^{-4}$ & $10^{-5}$ \\
Batch & effective 16 & 6 rollouts $\times$ 4 prompts \\
Length & 4,096 tokens, loss on assistant turns & 2,048 new tokens (curriculum: 768) \\
Schedule & 2 epochs over 353 trajectories & 60 steps \\
Other & gradient checkpointing & KL 0.02, temperature 1.0, on-policy \\
Seed & 0 & 0, 1, 2 (curriculum: 0) \\
\bottomrule
\end{tabular}
\end{table}

\paragraph{Comparator, sandbox and reward choice.}
The match $\equiv$ compares result rows as multisets, ordered when the reference is ordered; columns may be permuted and floats are rounded to 4 decimals. Execution is read-only, one statement, no \texttt{ATTACH}, a 5\,s timeout and at most 100k rows. The form of the CF reward was fixed before training by a pre-set rule: fewest correct queries rejected, then most GRPO groups with unequal rewards on 240 SFT rollouts.

\paragraph{System prompt} (verbatim; the user turn is \texttt{Database: <id>} and \texttt{Question: <question>}):
\begin{quote}\small\ttfamily
You are a careful data analyst working with a retail SQLite database.\\
Schema: \{SCHEMA\}\\
Business definitions: \{GLOSSARY\}\\
Use the run\_sql tool to explore the data and check your query before answering. Watch for duplicate rows, joins that multiply rows, missing business filters and superseded records. When you are done, reply with one short sentence and your final query in a \textasciigrave\textasciigrave\textasciigrave sql block. The final query must compute the answer from the tables; do not hard-code values.
\end{quote}

\paragraph{Engineering fixes.}
Before any reported run we fixed three issues that change results: AgentTune's multi-turn token stream dropped chat-template tokens between turns (importance ratio about $10^{-5}$), so we re-render the conversation and train strictly on-policy; a bf16 LoRA merge left 38--61\% of weights bit-identical to the base, so we merge in fp32 and store fp16; and an evaluation seed bug gave identical samples per prompt.

\paragraph{Compute.}
SFT: 72 minutes on an RTX 5070 Ti. GRPO main pair: about 7 hours on one L4 (about \$5). Curriculum pair: about 8.5 hours on one L4 (about \$6). Seeds 1--2: about 14 hours (about \$10). LeakySQL pilot: about 7 hours (about \$5). Test generation: about 13 minutes per model; Spider: 36 minutes for 4 models. The audit runs on a CPU in minutes.

\section{Additional results}
\label{app:results}

\paragraph{Pay rate per mutant type and published systems.}
Table~\ref{tab:mutants} breaks the audit down by mutant type. Table~\ref{tab:spider} grades eight published Spider systems; our small policies are luckier, with 18--28\% of their EX passes failing a suite database.
\begin{table}[ht]
\caption{\textbf{Pay rate per mutant type.} Share of audit mutants paid in full (\%). Set, $D_0$: set match on the original database; $\tilde R$: strict single-DB reward; $R$, blind: strict CF reward with blind copies; $n$: mutants.}
\label{tab:mutants}
\centering\small
\begin{tabular}{llrrrr}
\toprule
Mutant & Family & $n$ & Set, $D_0$ & $\tilde R$ & $R$, blind \\
\midrule
\texttt{all\_customers} & F1 & 40 & 40.0 & 40.0 & 40.0 \\
\texttt{no\_distinct} & F1 & 203 & 100.0 & 28.6 & 7.4 \\
\texttt{count\_orders} & F1 & 134 & 24.6 & 24.6 & 10.4 \\
\texttt{count\_rows} & F1 & 174 & 8.6 & 8.6 & 1.1 \\
\texttt{join\_fanout} & F2 & 208 & 30.3 & 30.3 & 20.2 \\
\texttt{missing\_refunds} & F3 & 77 & 54.5 & 54.5 & 37.7 \\
\texttt{no\_status\_filter} & F3 & 348 & 39.7 & 39.7 & 25.6 \\
\texttt{shipped\_only} & F3 & 77 & 36.4 & 36.4 & 20.8 \\
\texttt{item\_level\_avg} & F3 & 77 & 2.6 & 2.6 & 1.3 \\
\texttt{ignore\_adjustments} & F4 & 75 & 69.3 & 69.3 & 36.0 \\
\texttt{any\_row} & F4 & 217 & 25.8 & 25.8 & 16.6 \\
\texttt{first\_row} & F4 & 107 & 21.5 & 21.5 & 13.1 \\
\texttt{all\_tier\_rows} & F4 & 77 & 11.7 & 11.7 & 5.2 \\
\bottomrule
\end{tabular}

\end{table}
\begin{table}[ht]
\caption{\textbf{Published Spider systems pass test-suite-failing queries.} ESM+ predictions~\citep{ascoli2025etm} graded with our strict comparator. Lucky share: EX passes that fail a test-suite database (\%, Wilson 95\% CI). Lucky shares are upper bounds, because some suite failures are ties or ambiguous references.}
\label{tab:spider}
\centering\small
\begin{tabular}{lrrrr}
\toprule
System & EX (\%) & TS (\%) & Lucky / EX passes & Lucky share (\%) \\
\midrule
Claude & 80.6 & 68.6 & 124/833 & 14.9{\,[12.6, 17.5]} \\
C3 & 79.7 & 68.5 & 116/824 & 14.1{\,[11.9, 16.6]} \\
DIN & 81.5 & 72.2 & 96/843 & 11.4{\,[9.4, 13.7]} \\
GPT4Turbo & 72.3 & 64.7 & 79/748 & 10.6{\,[8.6, 13.0]} \\
DAIL & 82.7 & 75.3 & 76/855 & 8.9{\,[7.2, 11.0]} \\
RASAT+PICARD & 76.5 & 70.1 & 66/791 & 8.3{\,[6.6, 10.5]} \\
RESDSQL & 82.6 & 75.8 & 70/854 & 8.2{\,[6.5, 10.2]} \\
Graphix & 79.9 & 73.4 & 67/826 & 8.1{\,[6.4, 10.2]} \\
\bottomrule
\end{tabular}

\end{table}

\begin{figure}[ht]
\centering
\includegraphics[width=\linewidth]{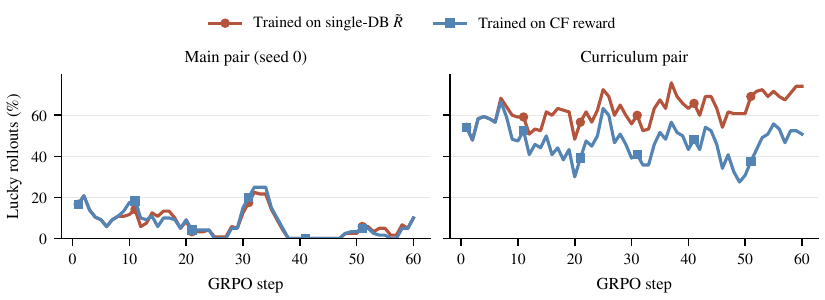}
\caption{\textbf{Under the single-DB reward, lucky rollouts grow only on hard prompts.} Share of rollouts per GRPO step that pass $D_0$ but fail a training copy (5-step trailing mean, one run each). \textbf{Left:} main pair, seed 0. \textbf{Right:} curriculum pair. Rust, circles: trained on $\tilde R$; blue, squares: trained on the CF reward.}
\label{fig:monitor}
\end{figure}

\paragraph{Monitoring the gap during training.}
Figure~\ref{fig:monitor} tracks the lucky share during training.
Share of rollouts per GRPO step that pass $D_0$ but fail a training copy (first 10 to last 10 steps): main pair 10.4\% to 7.5\% ($\tilde R$) and 13.3\% to 5.8\% ($R_{\text{mean}}$); curriculum pair 58.8\% to 72.9\% ($\tilde R$) and 52.9\% to 53.3\% ($R$). 
Replaying the logged rewards, $R_{\text{mean}}$ would have given a non-zero advantage to 28 of the main CF run's 240 groups against 26 for $\tilde R$; on the single-DB curriculum run, 105 against 53. 

\paragraph{SkyRL-SQL.} Grading the SkyRL-SQL set-match reward on copies removes 10.3 points of mutant pay (37.5\% to 27.2\%) and a strict comparator removes 10.4 more.

\paragraph{Bug visibility and database size.}
The share of mutants that pass the single-DB check is 29.2\% at 50 customers, 0.8\% at 500 and 0.24\% at 5,000.

\paragraph{Other results, all null or negative.}
DPO from SFT rollouts with single-DB labels reaches 9.7\% verified and with CF labels 8.1\% ($-1.6$ [$-2.6$, $-0.6$]). Partial-credit rewards add training signal but pay wrong answers more (mean pay on mutants 0.65--0.66 against 0.30 for $\tilde R$); we did not train on them. Majority voting over 8 samples raises GRPO-B from 13.9\% to 16.9\%, and voting on all 5 test databases adds nothing over voting on 1. An extra copy that moves orders to year boundaries lowers date bugs paid from 92.1\% to 32.9\%, with 0 of 1,554 correct queries rejected (\texttt{e12\_period\_copies.json}); it rejects at most 2 verified answers of either GRPO policy, so its effect on accuracy is under 1 point.

\paragraph{Reference point.}
Cohere Command A~\citep{cohere2025commandaenterprisereadylarge}, prompted once without tools, verifies 81.0\% [76.4, 84.8] of test items at \$4.24 per 1,000 queries; self-hosted GRPO-B costs \$0.061 per 1,000 queries on an L4 but verifies 12.7\%. Per verified answer, Tiny Aya is about 11$\times$ cheaper. The comparison is not like for like: one turn without tools against 5 tool steps and 8 samples.

\paragraph{Training seeds.}
\label{app:seeds}
Table~\ref{tab:seeds} gives every seed. SFT, the shared initialization, verifies 9.4\%.
\begin{table}[ht]
\caption{\textbf{Per-seed results.} Verified accuracy on the test set (\%) of arms B ($\tilde R$) and C (CF reward), and differences in points with question-bootstrap 95\% intervals; Spider: test-suite accuracy. Seed level: mean and Student-$t$ 95\% interval over the three per-seed differences.}
\label{tab:seeds}
\centering\small\setlength{\tabcolsep}{3pt}
\begin{tabular}{lrrrrrr}
\toprule
Seed & B & C & C$-$B & B$-$SFT & C$-$SFT & Spider C$-$B \\
\midrule
0 & 12.7 & 9.0 & $-3.7${\,[$-5.3$, $-2.3$]} & $+3.2${\,[$+1.8$, $+4.8$]} & $-0.5${\,[$-1.5$, $+0.6$]} & $+1.9${\,[$+0.6$, $+3.3$]} \\
1 & 7.6 & 8.8 & $+1.2${\,[$0.0$, $+2.6$]} & $-1.9${\,[$-3.0$, $-0.8$]} & $-0.6${\,[$-1.7$, $+0.5$]} & $-0.5${\,[$-1.8$, $+0.7$]} \\
2 & 14.7 & 7.6 & $-7.2${\,[$-10.0$, $-4.6$]} & $+5.3${\,[$+3.3$, $+7.4$]} & $-1.9${\,[$-3.2$, $-0.6$]} & $+3.3${\,[$+1.8$, $+4.7$]} \\
Pooled & 11.7 & 8.4 & $-3.2${\,[$-4.7$, $-1.8$]} & $+2.2${\,[$+1.0$, $+3.5$]} & $-1.0${\,[$-2.0$, $-0.1$]} & $+1.6${\,[$+0.8$, $+2.3$]} \\
Seed level & & & $-3.2${\,[$-13.7$, $+7.3$]} & $+2.2${\,[$-6.9$, $+11.4$]} & $-1.0${\,[$-2.9$, $+0.9$]} & $+1.6${\,[$-3.2$, $+6.4$]} \\
\bottomrule
\end{tabular}

\end{table}

At 80\% power, one seed detects a C$-$B difference of 1.9--3.9 points and the three-seed pool 2.1 points, while the per-seed differences have a standard deviation of 4.2 points: question-level precision is not the bottleneck, training variance is.

\paragraph{LeakySQL pilot.}
\label{app:leaky}
A model organism of reward hacking~\citep{macdiarmid2025naturalemergentmisalignmentreward}: every database gets a table \texttt{saved\_reports} holding the original database's reference answers for half of its questions, and every copy gets the same, now stale, table, so reading it pays under $\tilde R$ and fails any answer-changing copy. Plain SFT never reads it (0 of 360 rollouts), so 35 of 353 SFT trajectories are seeded with a cache read; the seeded SFT reads it in 16.9\% of test samples. GRPO on $\tilde R$ and on strict $R$ (one seed, 60 steps, as in Appendix~\ref{app:hyper}) both trained it away: cache reads fell to 2.1\% and 1.5\% of test samples (C$-$B $-0.6$ [$-1.4$, $+0.2$]), and in training from 10.4\% to 1.7\% and from 9.6\% to 0.0\% of rollouts. $\tilde R$ paid 39 of 60 training cache reads, but at a mean reward of 0.65 against 0.88 for other rollouts the hack never out-earned honest SQL; $R$ paid 0 of 26. Verified accuracy rose from 8.1\% (seeded SFT) to 15.4\% under $\tilde R$ and 12.9\% under $R$ (C$-$B $-2.5$ [$-4.0$, $-1.1$]). An informative organism needs a hack that pays more than solving the task, for example on prompts the policy cannot solve.

\section{Licenses, funding and use of LLMs}
\label{app:licenses}

Our code is Apache-2.0. Tiny Aya Global and our adapters are CC-BY-NC-4.0 (gated by the base model). Our generated data and trajectories are CC-BY-NC-4.0; released Spider trajectories contain Spider rows and fall under Spider's CC BY-SA 4.0. Spider and the test-suite databases are not redistributed. ESM+ predictions are used for analysis only and we store question indices, not predictions. \emph{Funding:} this work was self-funded and received no external funding. \emph{Use of LLMs:} we used LLMs as an aid; the authors are responsible for all content.

\end{document}